\documentclass[10pt,twocolumn,letterpaper]{article}

\usepackage{cvpr}              

\usepackage{times}
\usepackage{epsfig}
\usepackage{graphicx}
\usepackage{amsmath}
\usepackage{amssymb}
\usepackage{animate}
\usepackage{booktabs}
\usepackage{multirow}
\usepackage{multicol}
\usepackage[accsupp]{axessibility}
\usepackage[dvipsnames]{xcolor}

\definecolor{cvprblue}{rgb}{0.21,0.49,0.74}
\usepackage[pagebackref,breaklinks,colorlinks,citecolor=cvprblue]{hyperref}

\def\paperID{9162} 
\def\confName{CVPR}
\def\confYear{2024}

\title{Knowledge-Enhanced Dual-stream Zero-shot Composed Image Retrieval}

\author{Yucheng Suo \quad
    Fan Ma \quad
    Linchao Zhu$^\dag$ \quad
    Yi Yang \\
    ReLER, CCAI, Zhejiang University, China \\
    {\small$^\dag$ Corresponding author} \\
}

\begin{document}
\maketitle
\begin{abstract}
We study the zero-shot Composed Image Retrieval (ZS-CIR) task, which is to retrieve the target image given a reference image and a description without training on the triplet datasets. Previous works generate pseudo-word tokens by projecting the reference image features to the text embedding space. However, they focus on the global visual representation, ignoring the representation of detailed attributes, \eg, color, object number and layout. 
To address this challenge, we propose a Knowledge-Enhanced Dual-stream zero-shot composed image retrieval framework (KEDs). 
KEDs implicitly models the attributes of the reference images by incorporating a database. The database enriches the pseudo-word tokens by providing relevant images and captions, emphasizing shared attribute information in various aspects. In this way, KEDs recognizes the reference image from diverse perspectives.
Moreover, KEDs adopts an extra stream that aligns pseudo-word tokens with textual concepts, leveraging pseudo-triplets mined from image-text pairs. The pseudo-word tokens generated in this stream are explicitly aligned with fine-grained semantics in the text embedding space. Extensive experiments on widely used benchmarks, \ie ImageNet-R, COCO object, Fashion-IQ and CIRR, show that KEDs outperforms previous zero-shot composed image retrieval methods. Code is available at \url{https://github.com/suoych/KEDs}.
\end{abstract}    
\section{Introduction}
Composed Image Retrieval (CIR) is a task first introduced by Vo \etal \cite{vo2019composing},  which involves retrieving the target image given a reference image and a modification description. Different from traditional image-based retrieval \cite{datta2008image,rui1999image,gordo2016deep} or text-based retrieval \cite{rasiwasia2010new,wang2017adversarial} tasks, composed image retrieval requires the model to interpret both visual and text modality information. In practical scenarios, CIR allows users to specify fine-grained styles and content details in the queries, enabling flexibility and customization.

\begin{figure}[t]
\begin{center}
\includegraphics[width=1.0\linewidth]{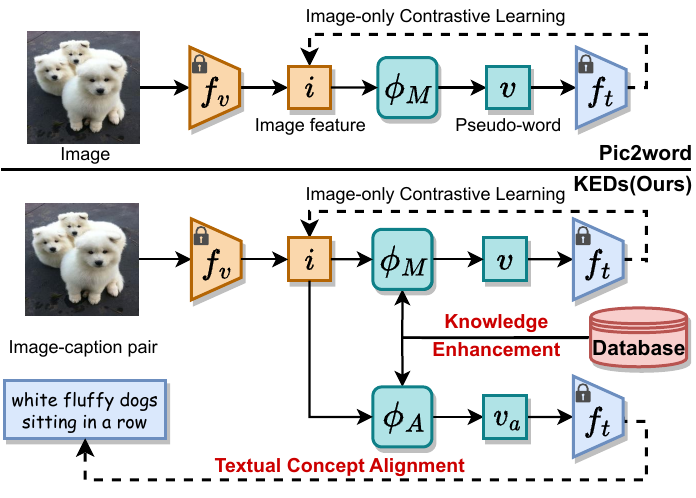}
\end{center}
\vspace{-.25in}
   \caption{\textbf{Comparison between existing methods and KEDs.} Pic2word \cite{saito2023pic2word} learns the mapping network using image-only contrastive learning and generates pseudo work $\phi_M$ token $v$. We propose to augment the pseudo-word token with external knowledge. In addition, we introduce an extra branch $\phi_A$ sharing architecture with $\phi_M$ for textual concept alignment. Note that $f_v$ and $f_t$ indicate frozen CLIP visual encoder and text encoder respectively.
   }
\label{fig:comparison}
\vspace{-.2in}
\end{figure}

With the rise of image-text pre-training models like CLIP \cite{radford2021learning}, significant improvements are achieved in the composed image retrieval field \cite{baldrati2022effective,delmas2022artemis,liu2021image}. Previous supervised methods work on designing networks to generate compositional features based on the reference image and text. Training and evaluation are conducted on the triplet datasets in various domains, \eg Fashion-IQ \cite{wu2021fashion} on the fashion clothes domain, CIRR \cite{liu2021image} on real-world images. Despite the impressive performance achieved by these supervised methods, two major limitations remain. Firstly, these approaches require extensive triplet data annotations, which are labor-intensive and time-consuming. Secondly, supervised methods train a tailored model for each dataset, thereby reducing the flexibility and generalization ability.

Recent studies \cite{baldrati2023zero, saito2023pic2word} introduce zero-shot approaches for composed image retrieval to address the above limitations. Pic2word is the pioneering work that learns a mapping layer that projects the image features into the text embedding space. This is achieved by self-supervised learning using a contrastive loss between the mapped features and original image features as the training objective. A frozen CLIP text encoder then generates the hybrid feature for inference. This image-only training paradigm learns a single model capable of datasets in various domains without triplet data. SEARLE \cite{baldrati2023zero} is another work that uses a large language model to generate additional descriptions based on the object class, enhancing the alignment between mapped image features and class semantics. However, these approaches share a common limitation: they directly map the overall image feature to the text embedding space, overlooking the detailed attribute information. This reduces the retrieval accuracy since the text in composed retrieval triplets only describes the object differences between target images and reference images.

To address the aforementioned issues, we propose a Knowledge-Enhanced Dual Stream zero-shot composed image retrieval framework (KEDs). 
First of all, KEDs incorporates a Bi-modality Knowledge-guided Projection network (BKP) that generates pseudo-word tokens based on external knowledge. Specifically, BKP incorporates a database to provide relevant images and captions to generate comprehensive pseudo-word tokens. The advantage is that the retrieved captions and images enrich the pseudo-word tokens with shared attribute information, \eg, object number and layout. This approach is akin to an ``open-book exam'', where the database serves as a reference sheet to better identify the reference image. We simply construct the database by random sampling a portion of the image-text pairs from the training dataset.

Image-only contrastive training does not align the pseudo-word tokens with real text concepts, bringing difficulty in composing reference images with text during inference. Therefore, we introduce an extra stream to generate pseudo-word tokens aligned with textual semantics. This stream is trained on pseudo-triplets mined from image-text pairs. This approach facilitates the interaction between pseudo-word tokens and diverse text during inference while maintaining the object semantics. 
Note that the pseudo-triplet mining process does not require generating extra data by external models.  
During inference, KEDs combines the output of the two streams, allowing for controlled alignment with specific modalities.

To evaluate the effectiveness of KEDs, we conduct experiments across four datasets, \ie ImageNet-R \cite{hendrycks2021many,deng2009imagenet}, COCO \cite{lin2014microsoft}, Fashion-IQ \cite{wu2021fashion}, CIRR \cite{liu2021image}. The four datasets test KEDs on different aspects of compositional ability, showcasing the generalization ability. The result indicates that KEDs surpasses all previous methods, especially in the ImageNet-R domain conversion task, achieving a remarkable boost of  $7.9\%$ in Recall@10 and $12.2\%$ in Recall@50 on average. Additionally, ablation studies are also provided to inspect the details of the method design. Overall, our contributions can be concluded as follows:
\begin{itemize}
    \item We propose a Knowledge-Enhanced Dual Stream framework (KEDs) for zero-shot composed image retrieval, where an external database enriches the mapping network with knowledge, enhancing the generalization ability.
    \item A new textual concept alignment training paradigm utilizing pseudo-triplets mined from the image-text pairs, ensuring semantic alignment between the mapped visual features and rich semantics.
    \item Extensive experiments on four datasets demonstrate the effectiveness of the proposed framework. 
\end{itemize}

\section{Related Work} \label{related}
\subsection{Composed Image Retrieval}
Composed Image Retrieval (CIR) is a compositional task first introduced by Vo \etal \cite{vo2019composing}, which aims to retrieve a target image given a reference image and a modification description \cite{yu2020curlingnet}. A critical aspect of the task is the extraction and combination of information from both reference images and text \cite{wu2023few,chen2020learning}. Current supervised methods train a cross-modal network using a fusion paradigm, which learns a joint embedding combining image and text. Representative methods include CoSMo \cite{lee2021cosmo}, DCNet \cite{kim2021dual}, ARTEMIS \cite{delmas2022artemis}, CLIP4Cir \cite{baldrati2022effective}, \etc. These supervised methods are trained on various labeled triplet data benchmarks including Fashion-IQ \cite{wu2021fashion}, CIRR \cite{liu2021image}, \etc. 

Considering the labor-intensive process of obtaining triplet data, Saito \etal \cite{saito2023pic2word} introduce a new setting to train a network without triplet data under a zero-shot setting. Their pioneer work Pic2word learns a mapping network that projects the image feature to the text embedding space, achieving impressive performance over datasets like CIRR and Fashion-IQ. Baldrati \etal design a text inversion network distilling the mapping network for better alignment with nouns \cite{baldrati2023zero}. Another line of work focuses on generating extra triplet data using generative models \cite{gu2023compodiff,liu2023zero}. In this paper, we propose a method trained on image-text pairs since image-text data pairs are easy to acquire and contain pair-matching prior information. 

\begin{figure*}[t]
\begin{center}
\includegraphics[width=1.0\linewidth]{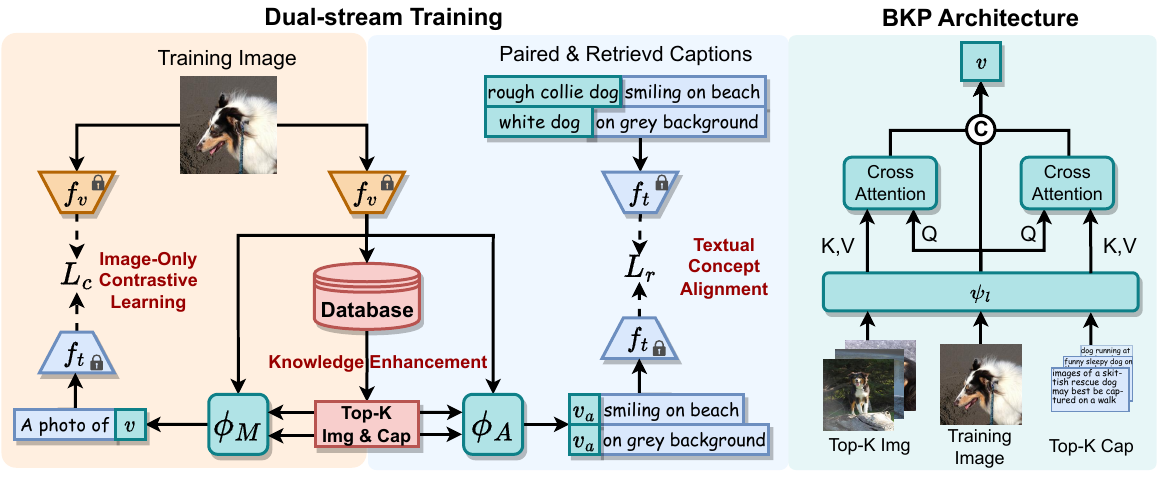}
\end{center}
\vspace{-.25in}
   \caption{\textbf{Overall framework of KEDs.} The left part of the figure represents the dual-stream training of KEDs, consisting of the image-only contrastive training (orange) and textual concept alignment branch (blue). The right part represents the architecture of the proposed Bi-modality Knowledge-guided projection.}
\label{fig:structure}
\vspace{-.2in}
\end{figure*}

\subsection{Knowledge Enhanced Methods}
Knowledge-enhanced methods are widely used in the natural language processing community \cite{guu2020retrieval,lewis2020retrieval}. The fundamental concept is to improve the performance by incorporating external knowledge \cite{xie2023ra}. In essence, knowledge-enhanced methods require the model to generate predictions considering prior knowledge \cite{iscen2023retrieval}. The advantage lies in their ability to inject interpretable world knowledge, particularly beneficial for knowledge-intensive tasks like open-domain question answering. With the rise of Large Language Models, knowledge-enhanced methods are used for comprehending long context and improving generation quality \cite{hofstatter2023fid,ram2023context,izacard2022few,liu2022relational}.

Having demonstrated success in NLP, the effectiveness of knowledge-enhanced methods also shows in other fields \cite{blattmann2022retrieval,chen2022re,goyal2022retrieval}. Recent works related to zero-shot recognition \cite{karthik2022kg, shen2022k, li2022clip, norelli2022asif, long2022retrieval, cheng2023hybrid}. For instance, K-LITE \cite{shen2022k} trains a vision-text model with expanded entities via retrieving words from Wordnet \cite{miller1995wordnet} or Wikitionary\cite{meyer2012wiktionary}. RA-CLIP \cite{xie2023ra} utilizes an extra feature database to enhance cross-model knowledge, achieving the inference process in an "open-book exam" style since the feature database can be considered as a "cheating sheet". RECO \cite{iscen2023retrieval} explores the retrieving modalities and fusion methods in a similar setting.
In this paper, we retrieve relevant images and captions to enrich the pseudo-word token semantics.

\subsection{Vision-language Pretraining}
Vision-language pre-training has been a long-standing research topic with wide real-world applications \cite{desai2021virtex,gomez2017self,zhou2020unified,hu2016natural,jia2021scaling,ma2023temporal,ma2023vista}. CLIP \cite{radford2021learning} is a representative work using image-text pairs to train the visual encoder and text encoder in a contrastive manner, achieving remarkable zero-shot performance on downstream tasks, including classification and image-text retrieval. CLIP lays the foundation of later vision-text pertaining models such as CoCa \cite{yu2022coca}, BLIP \cite{li2022blip,li2023blip}, ALBEF \cite{li2021align}, Flamingo \cite{alayrac2022flamingo}, \etc.

Researchers have explored the strong zero-shot ability of CLIP in open-domain tasks like open-set recognition \cite{conde2021clip, zhang2022tip,esmaeilpour2022zero,he2023open,li2023clip}, open-vocabulary detection \cite{minderer2022simple,zhang2023simple,du2022learning,shi2023edadet,huang2022robust} and segmentation \cite{jeong2023winclip,liang2023open,xu2023side,hu2023suppressing}. Cohen \etal propose PALAVRA \cite{cohen2022my}, a learning paradigm for personalized concepts. PALAVRA learns mapping networks to project image embeddings to the text embedding space or project text embeddings to the image embedding space. This self-supervised method also appears in recent zero-shot papers \cite{li2023decap,saito2023pic2word,baldrati2023zero,gal2022image,tewel2023key}. In this paper, our goal is also to learn a mapping network but with an extra objective.
\section{Method} \label{Method}
In this section, we describe the proposed method in detail. The overall pipeline of KEDs is illustrated in Figure~\ref{fig:structure}. We first introduce the preliminaries in Section~\ref{PRE}. Then in Section~\ref{RAM}, we introduce the Bi-modality knowledge-enhanced projection. The dual-stream alignment training paradigm is discussed in section~\ref{KEPTA}. Additionally, We introduce the inference process of KEDs in Section~\ref{HI}.
\subsection{Preliminaries}\label{PRE}
In this work, all retrieval processes are accomplished via Contrastive Language Image Pre-training (CLIP). CLIP consists of a visual encoder $f_v$ and a text encoder $f_t$ trained on image-text pairs with a contrastive objective. Specifically, given an image $\boldsymbol{I}$ and corresponding caption $\boldsymbol{T}$, the visual encoder extracts visual feature $\boldsymbol{i} = f_v(\boldsymbol{I}) \in \mathbb{R}^d$, and the text encoder extracts the overall caption feature $\boldsymbol{t} = f_t(\boldsymbol{T})\in \mathbb{R}^d$. To align the image-text pair in a contrastive manner, CLIP calculates a symmetric cross-entropy loss $\mathcal{L}_c$ \cite{sohn2016improved} which can be formulated as:
\begin{align}
\mathcal{L}_{i2t} &=  - \frac{1}{ |\mathcal{B}| }\sum_{ n \in \mathcal{B} } \log \frac{ \exp(\tau \boldsymbol{i}_{n}^T \boldsymbol{t}_n)  }{\sum_{ m \in \mathcal{B}}  \exp(\tau \boldsymbol{i}_{n}^T \boldsymbol{t}_{m})},\\
\mathcal{L}_{t2i} &=  - \frac{1}{ |\mathcal{B}| }\sum_{ n \in \mathcal{B} } \log \frac{ \exp(\tau \boldsymbol{t}_{n}^T \boldsymbol{i}_n)  }{\sum_{ m \in \mathcal{B}}  \exp(\tau \boldsymbol{t}_{n}^T \boldsymbol{i}_{m})},\\
\mathcal{L}_c &= \mathcal{L}_{i2t} + \mathcal{L}_{t2i}. 
\end{align}
Note that the image features and the text features are normalized before calculating the contrastive loss.
\subsection{Bi-modality Knowledge-guided Projection}\label{RAM}
To train a model for zero-shot composed image retrieval without triplet datasets, we follow previous work \cite{saito2023pic2word} to learn a mapping network that projects the image feature into the text embedding space, forming a pseudo-word token. As shown in Figure~\ref{fig:structure}, we prompt the pseudo-word token $\boldsymbol{v}$ with the text "A photo of" and encode with a frozen CLIP model, the generated feature is used for calculating the contrastive loss $\mathcal{L}_c$ with the image feature $\boldsymbol{i}$. However, this image-only training process focuses on the global image representation, neglecting detailed attribute information. 

To overcome this issue, we propose a Bi-modality Knowledge-guided Projection network (BKP) $\phi_M$ to generate the pseudo-word token $\boldsymbol{v}$ extracting information from relevant images and captions. \\
\textbf{Top-K images and captions retrieval.} As illustrated in Figure ~\ref{fig:structure}, we provide bi-modality knowledge for each training image by retrieving Top-K image features $\{\boldsymbol{i}^r_k\}^K_{k=1}$ and caption features $\{\boldsymbol{t}^r_k\}^K_{k=1}$ from a database. The Top-K image and caption features provide context for the projection, augmenting the vanilla image feature mapping network with shared attribute information. The database is simply constructed by random sampling 0.5M image-caption pairs from the training set and encoded by a pre-trained CLIP model. BKP retrieves items from the database using the faiss library \cite{johnson2019billion}, ensuring training efficiency. For a thorough context comprehension, we set K to 16. \\
\textbf{Bi-modality context Fusion.} After obtaining bi-modality knowledge from the database, we fuse the reference image feature with the knowledge with a simple network.
The training image feature $\boldsymbol{i}$ and the retrieved features  $\{\boldsymbol{i}^r_k\}^K_{k=1}$ and $\{\boldsymbol{t}^r_k\}^K_{k=1}$ are first projected into a common feature space through a linear block $\psi_l$ to bridge the modality gap. Then the training image feature $\boldsymbol{i}$ is used as the query to interact with the retrieved image features and caption features via two cross-attention blocks respectively to facilitate comprehension of context. The final output $\boldsymbol{v} \in \mathbb{R}^{3\times d}$ is the concatenation of three tokens, \ie, a mapped image feature token $\hat{\boldsymbol{i}} = \psi_l(\boldsymbol{i})$ and two context-aware mapped tokens $\boldsymbol{v}_i \in \mathbb{R}^d, \boldsymbol{v}_c \in \mathbb{R}^d$. The procedure can be formulated as:
\begin{align}
\boldsymbol{v}_i &= \texttt{CrossAttn}(\hat{\boldsymbol{i}}, \psi_l(\{\boldsymbol{i}^r_k\}^K_{k=1}),  \psi_l(\{\boldsymbol{i}^r_k\}^K_{k=1})), \\
\boldsymbol{v}_c &= \texttt{CrossAttn}(\hat{\boldsymbol{i}},  \psi_l(\{\boldsymbol{t}^r_k\}^K_{k=1}),  \psi_l(\{\boldsymbol{t}^r_k\}^K_{k=1})),\\
\boldsymbol{v} &= \texttt{concat}(\hat{\boldsymbol{i}},\boldsymbol{v}_i,\boldsymbol{v}_c). 
\end{align}

BKP encodes the bi-modality context together with the reference images, thereby generating pseudo-word tokens with comprehensive attribute information.
\begin{table*}[t]
  \centering \scalebox{1}{
  \begin{tabular}{lcrr|rr|rr|rr|rr} 
  \toprule
  \multicolumn{2}{c}{}     &   \multicolumn{2}{c}{Cartoon} & \multicolumn{2}{c}{Origami}& \multicolumn{2}{c}{Toy} & \multicolumn{2}{c}{Sculpture}& \multicolumn{2}{c}{Average}\\
  \cmidrule(lr){3-4}
  \cmidrule(lr){5-6}
  \cmidrule(lr){7-8}
  \cmidrule(lr){9-10}
  \cmidrule(lr){11-12}
  \multicolumn{1}{l}{Supervision} & \multicolumn{1}{l}{Methods} & R10 & R50& R10 & R50& R10 & R50& R10 & R50& R10 & R50\\  \midrule
\multirow{5}{*}{{\textsc{Zero-shot}}}
  & Image-only  &0.3&4.5&0.2&1.8&0.6&5.7&0.3&	4.0            &0.4&4.0\\
  &  Text-only  &0.2&1.1&0.8&3.7&0.8&2.4&0.4&	2.0&0.5&2.3      \\
  &  Image$+$Text&2.2&13.3&2.0&10.3&1.2&9.7&1.6&	11.6&1.7&11.2                    \\
& Pic2word  &8.0&21.9&13.5&25.6&8.7&21.6&10.0&23.8&10.1&23.2 \\
& \textbf{KEDs}
&\textbf{14.8}&\textbf{34.2}&\textbf{23.5}&\textbf{34.8}&\textbf{16.5}&\textbf{36.3}&\textbf{17.4}&\textbf{36.4}&\textbf{18.0}&\textbf{35.4}              \\
  \cmidrule{1-12}
\multicolumn{1}{c}{CIRR}  &Combiner~\cite{baldrati2022effective} &6.1 &14.8&10.5&21.3&7.0&17.7&8.5&20.4&8.0&18.5\\
\multicolumn{1}{c}{Fashion-IQ}&Combiner~\cite{baldrati2022effective} &	6.0&16.9&7.6&20.2&2.7&10.9&8.0&21.6&6.0&17.4\\
\bottomrule
  \end{tabular}}
  \vspace{-1mm}
  \caption{\textbf{Results of the domain conversion experiment using ImageNet-R.} Our method achieves state-of-the-art result.}
  \label{tab:imgnet}
  \vspace{-1mm}
\end{table*}

\begin{figure}[t]
\begin{center}
\includegraphics[width=1.0\linewidth]{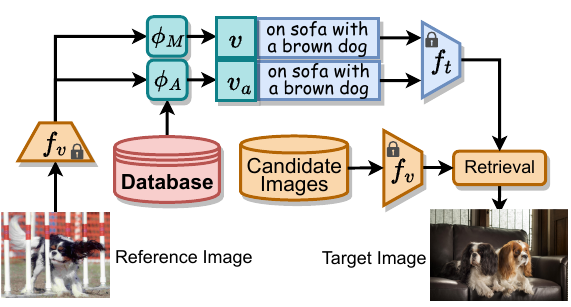}
\end{center}
\vspace{-.2in}
   \caption{A simple illustration of the inference process of KEDs.
   }
\label{fig:inference}
\vspace{-.2in}
\end{figure}

\begin{table}[t]
  \centering 
  \scalebox{1}{
  \begin{tabular}{c|c|ccc} 
  \toprule
  \cmidrule{1-5}
  \multicolumn{1}{c}{Supervision} & \multicolumn{1}{c}{Methods} & R1& R5&R10     \\ 
  \midrule
   \multirow{5}{*}{{\textsc{Zero-shot}}} & Image-only    &8.6&15.4&18.9\\ 
   & Text-only   &6.1&15.7&23.5\\
   & Image$+$Text  &10.2&20.2&26.6\\
   & Pic2word  &11.5&24.8&33.4\\
   & \textbf{KEDs} &\textbf{12.0}&\textbf{26.0}&\textbf{34.9}\\
     \cmidrule{1-5}
  \multicolumn{1}{c|}{CIRR} &Combiner~\cite{baldrati2022effective}&9.9&22.8&32.2\\
  \multicolumn{1}{c|}{Fashion-IQ}&Combiner~\cite{baldrati2022effective}&13.2&27.1&35.2\\
  \bottomrule
  \end{tabular}}
   \vspace{-1mm}
  \caption{\textbf{Evaluation on COCO object composition task.}}
  \vspace{-.15in}
  \label{tab:coco}
\end{table}
\subsection{Dual-stream Semantic Alignment}\label{KEPTA}
Previous work \cite{saito2023pic2word} trains the mapping network through image-only contrastive training as illustrated in Figure ~\ref{fig:structure}. A limitation is that the pseudo-word tokens are not aligned with the textual concepts in the text embedding space, introducing challenges in composing images with text during inference. To this end, we introduce training an extra stream $\phi_A$ on Pseudo-triplets mined from image-caption pairs to generate pseudo word token $v_a$ aligned with semantics. \\
\textbf{Pseudo-Triplets Mining.} The salient object in an image is described by the subject noun phrase of the paired caption syntactically. Our intuition is that the pseudo-word token $\boldsymbol{v}_a$ should align with the corresponding subject phrase semantic. Therefore, we extract pseudo-triplets consisting of an image, a piece of context description, and a target caption. 
To mine a pseudo-triplet, we conduct dependency parsing on a caption $\boldsymbol{T}$ using spacy \cite{spacy} and replace the subject noun phrase with the pseudo-word token $\boldsymbol{v}_a$.\\
\textbf{Complementary Pseudo-Triplets.} Each caption describes the object from a perspective, combining diverse captions stimulates comprehension of the image from different angles. For example, a caption for a photo showing a white husky sitting on a couch can be ``Husky on the sofa'', while a supplementary caption could be ``A sleeping white dog'', the two captions together comprehensively describe the different attributes, \eg dog breed, color, action. Therefore, we retrieve two additional captions $\boldsymbol{T}_s$ and $\boldsymbol{T}^{'}_s$ for each caption, generating complementary triplets to enrich semantics.\\ 
\textbf{Semantic registration loss.}
To train an extra projection $\phi_A$ on the pseudo-triplets, we introduce a semantic registration loss.
The pseudo-word token $\boldsymbol{v}_a$ is injected into the tokenized embedding of the prompt $\boldsymbol{T}_o$ and encoded by a frozen CLIP text encoder $f_t$. The output feature $\hat{\boldsymbol{v_a}}$ is used for calculating a cosine embedding loss $\mathcal{L}_{cos}$ with the overall caption embedding $\boldsymbol{t} = f_t(\boldsymbol{T})$:
\begin{align}
\mathcal{L}_{cos} = 1 - \texttt{cos}(\hat{\boldsymbol{v}_a}, \boldsymbol{t}). 
\end{align}
For the complementary triplets, we calculate an averaged supplementary cosine embedding loss $\mathcal{L}_{sup}$ with the retrieved captions embedding $\boldsymbol{t}_s = f_t(\boldsymbol{T}_s)$ and $\boldsymbol{t}^{'}_s = f_t(\boldsymbol{T}^{'}_s)$. The overall semantic registration loss $\mathcal{L}_r$ is calculated by linearly combining $\mathcal{L}_{cos}$ and $\mathcal{L}_{sup}$: 
\begin{align}
\mathcal{L}_{sup} &= 1 - \frac{1}{2}(\texttt{cos}(\hat{\boldsymbol{v}_a}, \boldsymbol{t}_s) + \texttt{cos}(\hat{\boldsymbol{v}_a}, \boldsymbol{t}^{'}_s)),\\
\mathcal{L}_r &= \mathcal{L}_{cos} + \beta\times \mathcal{L}_{sup}. 
\end{align}

The semantic alignment stream explicitly matches the pseudo-word token $\boldsymbol{v_a}$ with the textual concepts in the text embedding space. The pseudo-triplet mining process only requires image-text pairs, which is not labor-intensive. Note that the $\phi_A$ uses the identical Bi-modality Knowledge-guided Projection architecture with $\phi_M$.

\subsection{Hybrid Inference}\label{HI}
During inference, KEDs generates hybrid features of image and text for retrieval. Specifically, as shown in Figure ~\ref{fig:inference}, KEDs generate the pseudo-word tokens $\boldsymbol{v}$ and $\boldsymbol{v}_a$ given the reference image, then replace the placeholder token in the tokenized text embedding. In this way, we generate the composed feature using a frozen CLIP text encoder. As introduced in section ~\ref{RAM} and ~\ref{KEPTA}, we use a dual-stream training projection network. During inference, the two streams generate two different composed features $\hat{\boldsymbol{v}}$ and $\hat{\boldsymbol{v}_a}$ through the text encoder, we conduct a simple yet effective linear combination to generate a hybrid feature $\boldsymbol{v}_h$ for retrieval:
\begin{align}
\boldsymbol{v}_h = \alpha \times \hat{\boldsymbol{v}} + (1-\alpha) \times \hat{\boldsymbol{v}_a}. 
\end{align}
$\boldsymbol{v}_h$ is used for calculating the similarity with the candidate image features. The image with the highest similarity is selected as the prediction. Under this zero-shot inference process, KEDs is capable of various compositional datasets.

\begin{table*}[t]
  \centering \scalebox{1.0}{
  \begin{tabular}{lcrr|rr|rr|rr} 
  \toprule
  \multicolumn{2}{c}{}     &   \multicolumn{2}{c}{Dress} & \multicolumn{2}{c}{Shirt}& \multicolumn{2}{c}{TopTee} & \multicolumn{2}{c}{Average}\\
   \cmidrule(lr){3-4}
  \cmidrule(lr){5-6}
  \cmidrule(lr){7-8}
  \cmidrule(lr){9-10}
  \multicolumn{1}{l}{Supervision} & \multicolumn{1}{c}{Methods} & R10 & R50& R10 & R50& R10 & R50& R10 & R50\\  \midrule
  \multirow{6}{*}{{\textsc{Zero-shot}}}
  & Image-only  &5.4&13.9&9.9&20.8&8.3&17.7&7.9&17.5\\
  &  Text-only  &13.6&29.7&18.9&31.8&19.3&37.0&17.3&32.9\\
  &  Image$+$Text& 16.3&33.6&21.0&34.5&22.2&39.0&19.8&35.7\\
  &  Pic2word &20.0&40.2&26.2&43.6&27.9&47.4&24.7&43.7\\
  &  SEARLE-XL &20.3&43.2&27.4&45.7&29.3&50.2&25.7&46.3\\
  &  \textbf{KEDs} &  \textbf{21.7}&\textbf{43.8}&\textbf{28.9}&\textbf{48.0}&\textbf{29.9}&\textbf{51.9}&\textbf{26.8}&\textbf{47.9}\\

\midrule
\multicolumn{1}{c}{CIRR}  &Combiner~\cite{baldrati2022effective} &17.2&37.9&23.7&39.4&24.1	&43.9&21.7&40.4 \\
\multicolumn{1}{c}{Fashion-IQ}&Combiner~\cite{baldrati2022effective}&30.3&54.5&37.2&55.8&39.2&61.3&35.6&57.2\\
\multicolumn{1}{c}{Fashion-IQ}&Combiner$^{*}$~\cite{baldrati2022effective}&31.6&56.7&36.4&58.0&38.2&62.4&35.4&59.0\\

\multicolumn{1}{c}{Fashion-IQ}& CIRPLANT~\cite{liu2021image}&17.5&40.4&17.5&38.8&21.6&45.4& 18.9 &41.5\\
\multicolumn{1}{c}{Fashion-IQ}&ALTEMIS~\cite{delmas2022artemis}&27.2&52.4&21.8&43.6&29.2&54.8&26.1&50.3\\
\multicolumn{1}{c}{Fashion-IQ}&MAAF~\cite{dodds2020modality}&23.8&48.6&21.3&44.2&27.9&53.6&24.3&48.8\\
\bottomrule
  \end{tabular}}
   \vspace{-1mm}
  \caption{\textbf{Results on Fashion-IQ validation set.} $^{*}$ is the result using ResNet50x4 backbone.}
  \vspace{-2mm}
  \label{tab:fashion}
\end{table*}

\begin{table}[t]
  \centering 
  \scalebox{0.9}{
  \begin{tabular}{c|c|cccc} 
  \toprule
  \multicolumn{1}{l}{Supervision} & \multicolumn{1}{c}{Methods} & R1& R5&R10&R50\\ 
  \midrule
   \multirow{6}{*}{{\textsc{Zero-shot}}} & Image-only    &7.4	&23.6&34.0&57.4\\
   & Text-only   &20.9&44.8&56.7&79.1\\
   & Image$+$Text  &12.4&36.2&49.1&78.2\\
   & Pic2word  &23.9&51.7&65.3&87.8\\
   & SEARLE-XL  &24.2&52.4&66.3&88.6\\
   & \textbf{KEDs} &\textbf{26.4}&\textbf{54.8}&\textbf{67.2}&\textbf{89.2}\\
     \cmidrule{1-6}
  \multicolumn{1}{c|}{CIRR} &Combiner~\cite{baldrati2022effective}&30.3&60.4&73.2&92.6\\
  \multicolumn{1}{c|}{Fashion-IQ}&Combiner~\cite{baldrati2022effective}&20.1&47.7&61.6&85.9\\
  
\multicolumn{1}{c|}{CIRR}&Combiner$^{*}$~\cite{baldrati2022effective}&33.6&65.4&77.4&95.2\\
\multicolumn{1}{c|}{CIRR}&TIRG~\cite{vo2019composing}&14.6&48.4&64.1&90.0\\
   \multicolumn{1}{c|}{CIRR}&ARTEMIS~\cite{delmas2022artemis}&17.0&46.1& 61.3&87.7\\
    \multicolumn{1}{c|}{CIRR}&CIRPLANT~\cite{liu2021image}&19.6&52.6&68.4&92.4\\
  \bottomrule
  \end{tabular}}
   \vspace{-1mm}
  \caption{\textbf{Evaluation on CIRR test set.} $^{*}$ is the result using ResNet50x4 backbone.}
  \vspace{-.2in}
  \label{tab:cirr_test}
\end{table}
\section{Experiments} \label{experiments}
In this section, we first introduce the benchmarks and experiment setup in section ~\ref{DS}. Then provide the implementation details and results on the different datasets in section ~\ref{ID} and ~\ref{QR}. We also provide a detailed ablation study and analysis of the experiment results in section ~\ref{AS}.
\subsection{Datasets and Setup} \label{DS}
KEDs is trained on the Conceptual Caption Three Million (CC3M) dataset \cite{sharma2018conceptual}. CC3M contains a wide variety of image-caption pairs and has no overlap with the evaluation datasets. For evaluation, we employ four datasets following previous work, \ie ImageNet-R \cite{hendrycks2021many,deng2009imagenet}, COCO \cite{lin2014microsoft}, Fashion-IQ \cite{wu2021fashion}, CIRR \cite{liu2021image}. The four datasets assess four types of composition ability individually:\\
\textbf{Domain conversion:} ImageNet-R \cite{hendrycks2021many,deng2009imagenet} is used in the domain conversion setup. Specifically, 16983 real images under 200 classes are required to be converted to four styles,\ie cartoon, origami, toy and sculpture. The correct target image should belong to the same class as the reference image while matching the domain description.\\ 
\textbf{Object composition:} COCO validation set \cite{lin2014microsoft} with 5000 images is employed for evaluating object composition. The reference images are constructed by cropping an object in the image according to the instance mask annotations and the goal is to retrieve the overall image.\\
\textbf{Scene manipulation:} In terms of the scene manipulation setup, we use the CIRR dataset \cite{liu2021image} consisting of crowd-sourced,
open-domain images with hand-written descriptions. Following previous works \cite{saito2023pic2word, baldrati2023zero}, we report the performance on the test split, while conducting ablation studies on the validation set.\\
\textbf{Fashion attribute manipulation:} Another widely used benchmark is Fashion-IQ \cite{wu2021fashion} designed for fashion images. The reference images indicate the type of clothes while the text describes the expected attributes. We report the results on the validation set.

\begin{figure}[t]
\begin{center}
\includegraphics[width=1.0\linewidth]{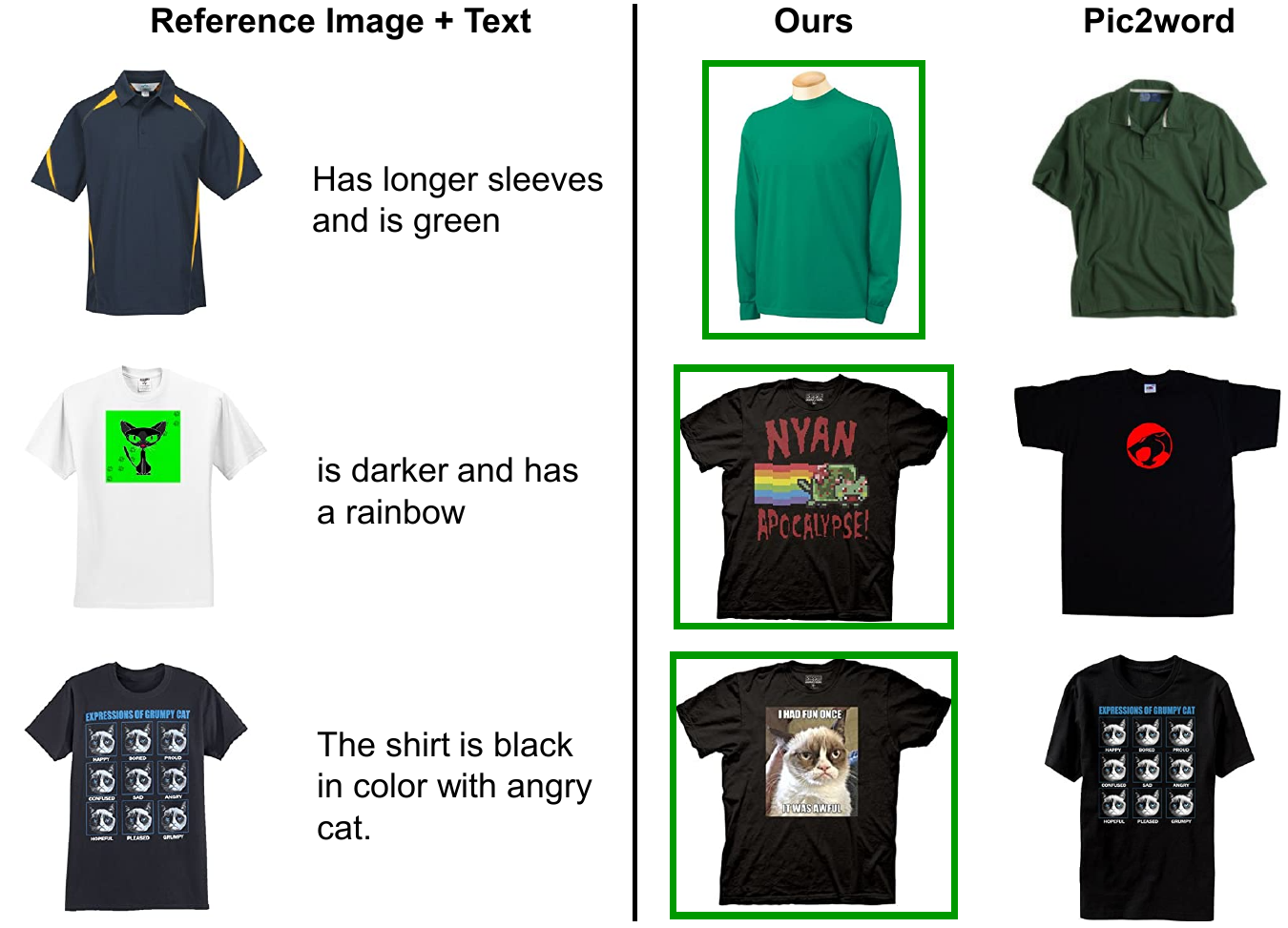}
\end{center}
\vspace{-.2in}
   \caption{Qualitative results on Fashion-IQ dataset. Images with green borders represent the ground truth.
   }
\label{fig:fashionvisual}
\vspace{-.1in}
\end{figure}
In section ~\ref{QR}, we quantitatively compare KEDs with the following baseline methods:\\
\textbf{Image-only} uses the similarity between the target image feature and reference image feature for retrieval. \\
\textbf{Text-only} retrieves the target image using the text features.\\
\textbf{Image+text} takes the average of the visual and text features to retrieve the target image.\\
\textbf{Pic2word} \cite{saito2023pic2word} is a method that learns a mapping network to project the image feature to the text embedding space. A frozen CLIP \cite{radford2021learning} text encoder fuses the mapped visual feature and text feature to retrieve the target image.\\
\textbf{SEARLE-XL} \cite{baldrati2023zero} generates descriptions using a large language model based on the class name and uses the descriptions to train a textual inversion network. The pseudo-word tokens are distilled from the inversion network.\\
\textbf{Supervised methods} \cite{baldrati2022effective,liu2021image,delmas2022artemis,dodds2020modality} are trained on the labeled triplet datasets including CIRR and Fashion-IQ. The performance is reported following previous work.
\subsection{Implementation Details} \label{ID}
We adopt the ViT-L/14 CLIP\cite{radford2021learning} backbone and use CC3M as the training dataset. The database is constructed by randomly selecting 0.5M image-text pairs and encoding the image and text by the pre-trained CLIP. We employ the GPU version faiss library \cite{johnson2019billion} for efficient real-time retrieval over the database. The Bi-modality Knowledge-guided projection module contains three layers of multi-head cross-attention with a hidden dim of 768. We use AdamW \cite{loshchilov2017decoupled} optimizer with a learning rate of $5e^{-5}$ and 0.1 weight decay. We conduct a linear warmup of 10000 steps and cosine learning rate decay to smooth the optimization. The model is trained for 30 epochs with a batch size of 512 on 8 RTX 4090 GPUs within one day.
\begin{figure}[t]
\begin{center}
\includegraphics[width=1.0\linewidth]{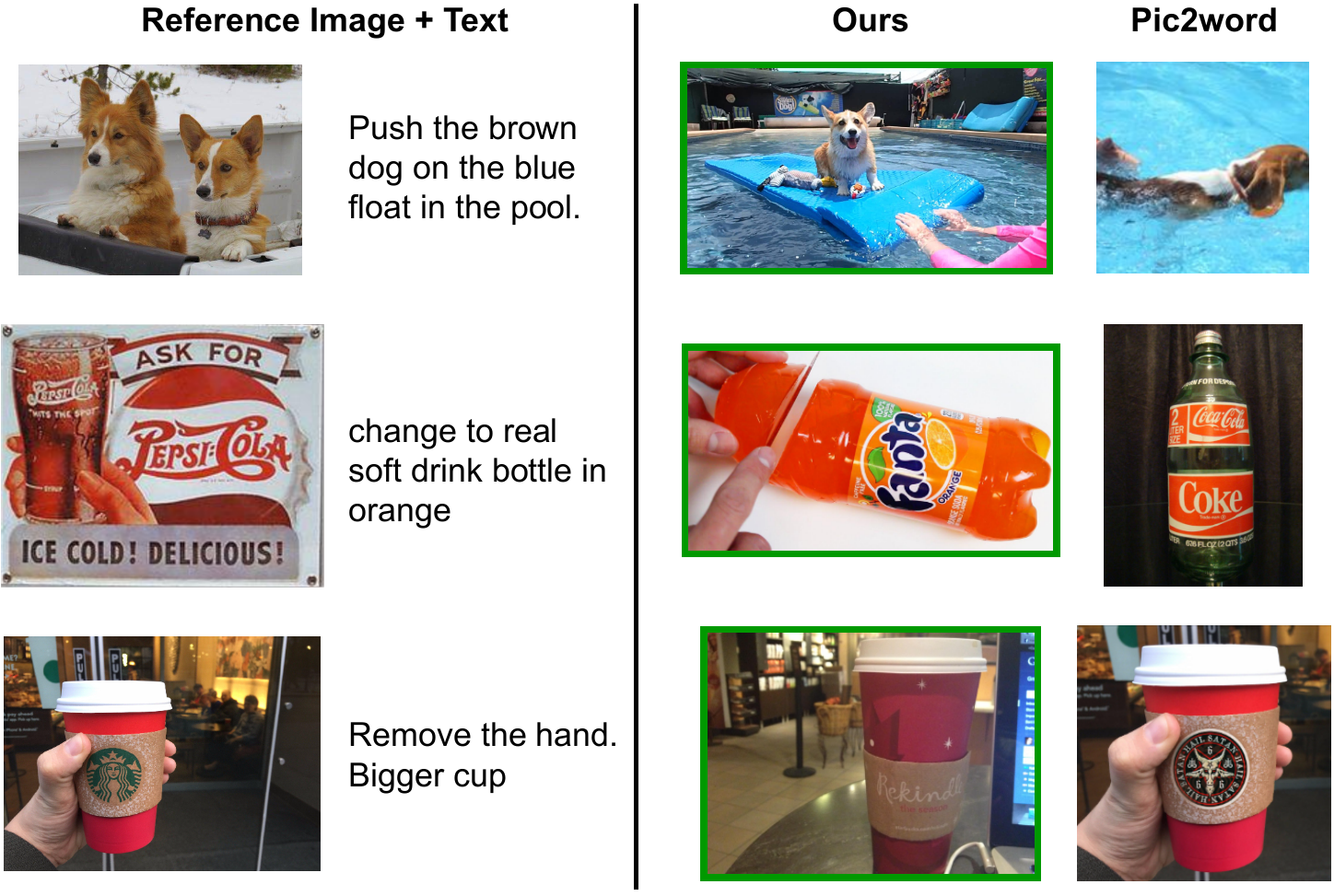}
\end{center}
\vspace{-.2in}
   \caption{Qualitative results on CIRR dataset. Images with green borders represent the ground truth.
   }
\label{fig:cirrvisual}
\vspace{.05in}
\end{figure}

\begin{table}[!t]
  \centering
  \Large
  \resizebox{\linewidth}{!}{ 
  \begin{tabular}{clccccc} 
  \toprule
  \multicolumn{1}{c}{} & \multicolumn{1}{c}{} & \multicolumn{3}{c}{CIRR} & \multicolumn{2}{c}{Fashion-IQ}\\
  \cmidrule(lr){3-5}
  \cmidrule{6-7}
  \multicolumn{1}{l}{Part} & \multicolumn{1}{l}{Method} & R1 & R5 & R10 & R10 & R50 \\ \midrule
 \multirow{2}{*}{All} & Pic2word &22.6&52.6&66.6 & 24.7 & 43.7 \\
  & \textbf{KEDs} & \textbf{27.3}&\textbf{56.4}&\textbf{69.2} & \textbf{26.8} & \textbf{47.9} \\ \midrule[.02em]
   \multirow{3}{*}{$\phi_M$} & w/o Top-K img &26.5&54.5&67.0 & 24.0 & 43.6 \\
  & w/o Top-K cap &26.3&54.5&66.7 & 22.5 & 42.2 \\
  & w/o linear & 21.7 & 49.0 & 60.9 & 15.9 & 31.6 \\ \midrule[.02em]

  \multirow{3}{*}{$\phi_A$}  
  & w/o $\phi_A$ &24.0&53.2&66.9 & 24.5 & 44.4 \\
  & w/o context &25.0&54.2&67.8 & 25.0 & 44.8 \\
  & w/o extra &27.2&56.0&68.5 & 26.0 & 46.3\\ \midrule[.02em]

  \multirow{2}{*}{DB}  & CIRR & 27.1 & 56.3 & 68.9 & 25.8 & 46.4 \\
  & Fashion-IQ & 26.7 & 55.7 & 68.5 & 26.0 & 46.4 \\
  
  \bottomrule
  \end{tabular}}
  \vspace{-.1in}
  \caption{Ablation studies on CIRR and FashionIQ validation sets. For FashionIQ, we consider the average recall. $\phi_M$ is the Bi-modality Knowledge-guided projection module, $\phi_A$ is the textual concept alignment branch, while DB represents the Database.} 
  \vspace{-.1in}
  \label{tab:cirr}
\end{table}

\begin{figure}[t]
\begin{center}
\includegraphics[width=1.0\linewidth]{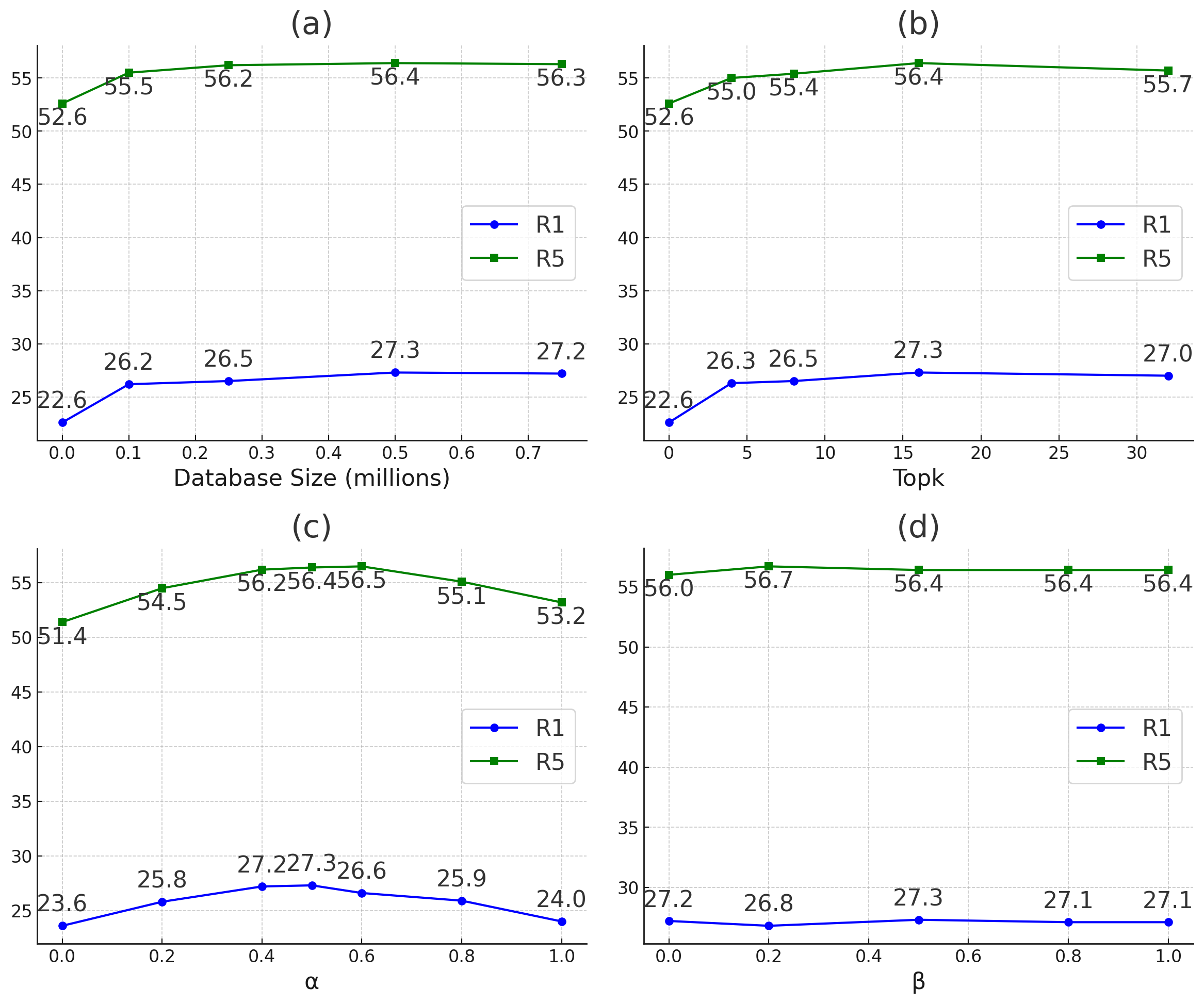}
\end{center}
\vspace{-.25in}
   \caption{Visualization of how (a) the item amount of the database, (b) the number of retrieved neighbors, (c) the weight $\alpha$ for mixture feature during inference, and (d) the loss weight $\beta$ influence the performance respectively on the CIRR validation set.
   }
\label{fig:dbtopk}
\vspace{-.15in}
\end{figure}

\begin{figure*}[t]
\begin{center}
\includegraphics[width=1.0\linewidth]{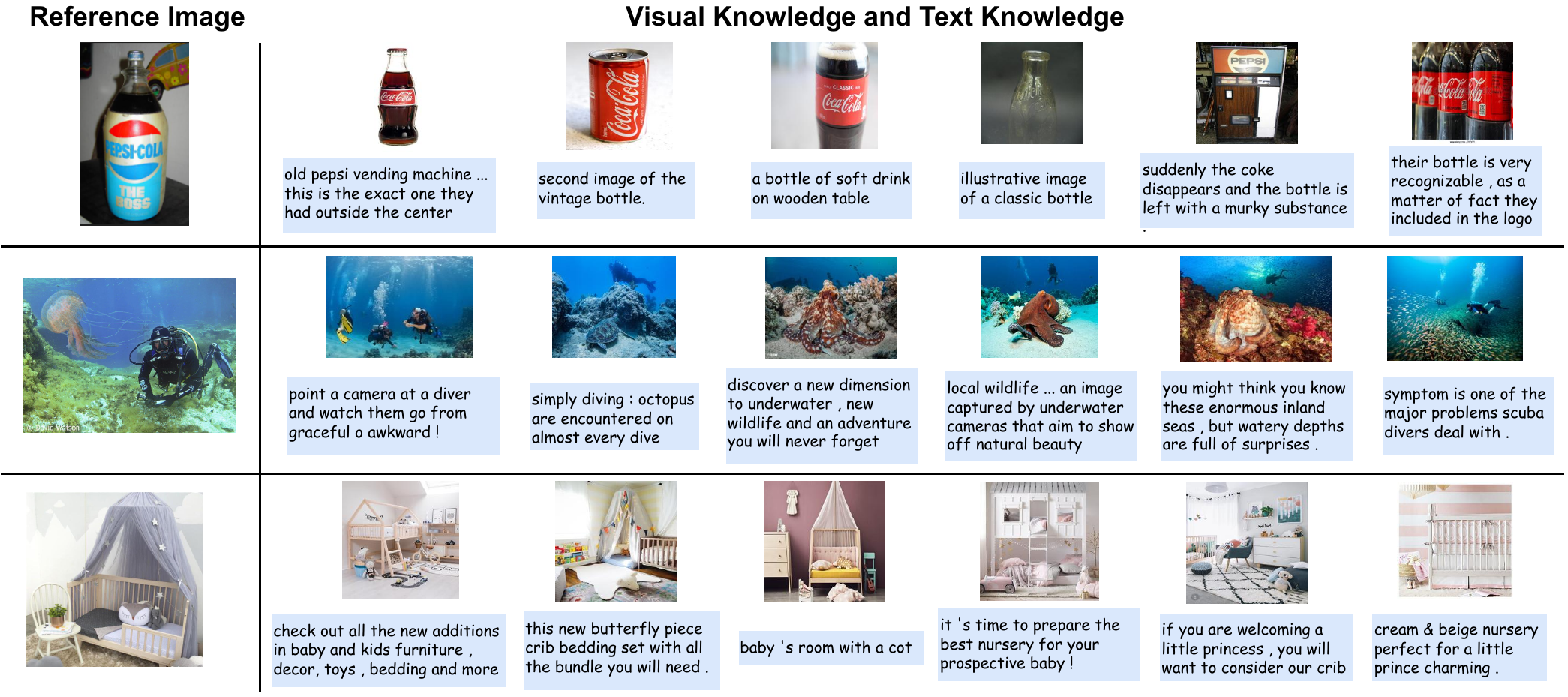}
\end{center}
\vspace{-.3in}
   \caption{Visualization of visual and text knowledge. The reference image is picked from the CIRR validation set and the visual and text knowledge is retrieved from CC3M.}
\label{fig:retrievalvisual}
\vspace{-.1in}
\end{figure*}

\subsection{Quantitative and Qualitative Results} \label{QR}
Table ~\ref{tab:imgnet} shows the performance of the ImageNet-R dataset, KEDs consistently outperforms all previous methods with a substantial margin, \ie, $+7.9\%$ Recall@10 and $+12.2\%$ Recall@50 on average. The boost is attributed to the alignment between pseudo-word tokens and object semantics. 

On the COCO object composition task, KEDs also outperforms previous methods ($+1.5\%$ Recall@10) as reported in Table ~\ref{tab:coco}. Notably, KEDs implicitly captures fine-grained object information, leveraging multi-modal knowledge from the database to recognize objects and deduce scene layouts from neighboring images.

In terms of the fashion attribute manipulation task reported in Table ~\ref{tab:fashion}, KEDs shows impressive results. For instance, there is a $1.5\%$ improvement on Recall@10 and $2.3\%$ on Recall@50 in the shirt domain compared with the previous state-of-the-art method SEARLE-XL \cite{baldrati2023zero}. The results demonstrate the generalization ability of KEDs. Even in a narrow domain, KEDs still captures the attributes of the reference image from neighbors.

Table ~\ref{tab:cirr_test} presents the result on the CIRR test set, we observe a $2.2\%$ and $2.4\%$ improvement in Recall@1 and Recall@5 respectively. KEDs achieves $26.4\%$ Recall@1, surpassing many supervised methods. The result reveals the real-world application potential of KEDs. Across these four datasets, KEDs consistently demonstrates its effectiveness in various compositional tasks.

We showcase qualitative example predictions by KEDs in Figure ~\ref{fig:cirrvisual} and Figure ~\ref{fig:fashionvisual}. Examples illustrate the ability of KEDs to comprehend the semantic meaning of modification description while preserving visual information from the reference image.

\subsection{Ablation Studies} \label{AS}
To verify the influence of the modules in KEDs, we conduct a detailed ablation study on several aspects, as reported in Table ~\ref{tab:cirr} and illustrated in Figure ~\ref{fig:dbtopk}.\\
\textbf{Effectiveness of the proposed modules.} We train KEDs with variants to evaluate the effectiveness of each module. Results are presented in Table ~\ref{tab:cirr}. For the BKP module $\phi_M$, we observe a consistent performance drop without the Top-K image or caption feature, indicating the importance of the knowledge modality. The performance also decreases when training without the shared linear block, emphasizing the importance of using a shared embedding space of $\phi_M$ to learn from external knowledge. Furthermore, the importance of the textual concept alignment branch $\phi_A$ is demonstrated by the performance boost. As mentioned in ~\ref{KEPTA}, the captions contain redundant semantics. Therefore the mapped features should specifically align with the subject phrases. This is achieved by offering contexts in $\phi_A$. When the context semantics are omitted, the performance dramatically decreases. Additionally, there is a slight performance drop without extra captions, aligning with the intuition that extra captions depict objects from diverse perspectives, aiding semantic alignment. 
We compare different weights $\alpha$ for mixing dual-stream features during inference, a consistent improvement over single-stream features is observed in Figure~\ref{fig:dbtopk} (c). In terms of the weights $\beta$ for the extra caption loss during training in Figure~\ref{fig:dbtopk} (d), results show that KEDs is robust to varying weights. \\
\textbf{Analysis of the Top-K values.} To find the optimal number of nearest neighbors to retrieve during training, we conduct experiments on the CIRR validation set and visualize the result in Figure~\ref{fig:dbtopk} (b). A small value of K provides limited information while a large value affects efficiency. To strike a balance, we set K to 16. We also visualize the retrieved items in the Figure~\ref{fig:retrievalvisual}. The retrieved images and captions are closely relevant to the reference image in certain attributes, assisting KEDs in extracting common information. The bi-modality knowledge in the relevant images and captions enriches the semantics of the generated pseudo-tokens.\\
\textbf{Design of the database.}
Figure~\ref{fig:dbtopk} (a) visualizes how the item amount in the database influences performance. While more items improve the performance, the benefit is marginal. Therefore we set the number of items to 0.5 Million image-caption pairs.
In the bottom part of Table ~\ref{tab:cirr}, we additionally test two variants of visual and text features in the database during inference, \ie substitute CC3M features with CIRR features or Fashion-IQ features. However, results indicate the CIRR features or Fashion-IQ features are not compatible with CC3M features. We believe the open-domain knowledge contributes to the zero-shot performance and the proposed Bi-modality Knowledge-guided Projection network learned the knowledge within the database.
\section{Conclusion} \label{conclusion}
In this paper, we propose a Knowledge Enhanced Dual Stream zero-shot composed image retrieval framework (KEDs) for Zero-shot Composed Image Retrieval. KEDs includes a Bi-modality Knowledge-guided Projection network. In particular, the network incorporates a database to provide relevant images and captions for the reference images. In this way, KEDs generates pseudo-word tokens with attribute information.
In addition, KEDs integrates an extra stream to generate pseudo-word tokens aligned with textual concepts in the text embedding space. During inference, we combine pseudo-word tokens from two streams for retrieval. Extensive experiments show that KEDs surpasses previous methods on four datasets including ImageNet-R, COCO, Fashion-IQ and CIRR. Future work may leverage a Large language model to generate detailed descriptions. 
\section{Acknowledgement}
This work is partially supported by Major program of the National Natural Science Foundation of China (Grant Number: T2293723). This work is also partially supported by the Fundamental Research Funds for the Central Universities (Grant Number: 226-2023-00126, 226-2022-00051) and China Postdoctoral Science Foundation (524000-X92302).

{
    \small
    \bibliographystyle{ieeenat_fullname}
    \bibliography{main}
}




\end{document}